\documentclass[letterpaper, 10 pt, conference]{ieeeconf}  

\IEEEoverridecommandlockouts                              
\usepackage{graphics} 
\usepackage{epsfig} 
\usepackage{mathptmx} 
\usepackage{times} 
\usepackage{amsmath} 
\usepackage{amssymb}  

\title{\LARGE \bf
Real-time regression analysis with deep convolutional neural networks
}

\author{E.~A.~Huerta$^{1,2}$, Daniel George$^{1,2}$,
Zhizhen Zhao$^{1,3}$, Gabrielle Allen$^{1,2}$\\ \\
$^{1}$NCSA, University of Illinois at Urbana-Champaign, Urbana, Illinois 61801, USA \\
$^{2}$Department of Astronomy, University of Illinois at Urbana-Champaign, Urbana, Illinois 61801, USA \\
$^{3}$Department of Electrical and Computing Engineering, University of Illinois at Urbana-Champaign, \\Urbana, Illinois 61801, USA \\
\{elihu, dgeorge5, zhizhenz, gdallen\}@illinois.edu\\
}

\vspace{-5mm}

\begin{document}

\maketitle
\thispagestyle{empty}
\pagestyle{empty}

\begin{abstract}

We discuss the development of novel deep learning algorithms to enable real-time regression analysis for time series data. We showcase the application of this new method with a timely case study, and then discuss the applicability of this approach to tackle similar challenges across science domains.

\end{abstract}

\section{Executive Summary}

In this article we discuss advances in deep learning research that we have pioneered and successfully applied for real-time \textit{classification} and \textit{regression} of time-series data both in simulated and in non-stationary, non-Gaussian data~\cite{DeepFiltering,DNNRealNoise}. We also describe a novel combination of deep learning and transfer learning that we have developed and applied for the classification and clustering of noise anomalies~\cite{DeepTransferLearning,DeepTransferNIPS}. We discuss the general applicability of these new methodologies across science domains to tackle computational grand challenges.

\section{Key Challenges}
The key challenges we address in this paper concern the application of deep neural networks (DNNs) to enable real-time \textit{classification} and \textit{regression} of time-series data that span a higher-dimensional parameter space, and which are embedded in non-Gaussian and non-stationary noise; and the combination of deep learning and transfer learning to develop deep neural networks for \textit{classification} and \textit{clustering} of noise anomalies using small training datasets. To highlight the relevance and timeliness of these research themes we consider two science cases: gravitational wave astrophysics and large scale astronomical surveys. 

The most sensitive algorithms that enabled the discovery of gravitational waves target only a 4-dimensional parameter space out of the 9-dimensional space that describes the gravitational wave sources available to the LIGO detectors. Limiting factors of these matched-filtering algorithms is their computational expense and lack of scalability for real-time regression analyses~\cite{owen:1999PhRvD..60b2002O,indik:2017,2016PhRvD..94b4012H,Smith:2016PhRvD}. This issue is exacerbated when matched-filtering is combined with fully Bayesian algorithms, leading to full production analyses that take from several hours to months using LIGO computing centers, and state-of-the-art high performance computing facilities such as the Blue Waters supercomputer~\cite{huerta:2017boss}. 

This computational grand challenge is ubiquitous across science domains. In the context of image processing,  the extraction of specific signatures from telescope images will become a major challenge once the Large Synoptic Survey Telescope---the most sensitive astronomical camera to take snapshots of the southern hemisphere---starts operating by the end of this decade. This astronomical facility will generate TB-size datasets on a nightly basis, releasing tens of thousands of images every minute that will contain an unprecedented amount of information of the nearby Universe. Key information embedded in these images will have to be processed in real-time to enable groundbreaking scientific discoveries. A task that is not feasible with existing algorithms.
 
\section{New Research Directions}

There are ongoing efforts to try to alleviate the lack of scalability of matched-filtering algorithms~\cite{indik:2017}. Other approaches involve the development of new signal processing techniques using machine learning~\cite{bambi:2012MNRAS,DBNN,jade1:2016,jade:2015CQGra,spy:2016arXiv,GravitySpy2,DeepTransferLearning,Denoising}. While these traditional machine learning techniques, including shallow artificial neural networks (ANNs), require ``handcrafted'' features extracted from the data as inputs rather than the raw noisy data itself, DNNs are capable of extracting these features automatically. 

In the context of image classification, we have applied deep learning for the classification of noise anomalies with spectrogram images as inputs to convolutional neural networks (CNNs)~\cite{GravitySpy,GravitySpy2,DeepTransferLearning} and unsupervised clustering of transients~\cite{DeepTransferLearning}. Using images as inputs is advantageous for two reasons: (i) there are well established architectures of 2D CNNs which have been shown to work (GoogLeNet~\cite{GoogLeNet}, VGG~\cite{VGG}, ResNet~\cite{ResNet}); and (ii) pre-trained weights are available for them, which can significantly speed up the training process via transfer learning while also providing higher accuracy even for small datasets~\cite{DeepTransferLearning}.

In the context of real-time regression, our new deep learning method performs very well when we consider a 2-dimensional parameter space. However, we want to explore its performance in higher-dimensional parameter space. A key feature that will prove critical in that scenario is the scalability of deep learning, i.e., all the intensive computation is diverted to the one-time training stage, after which the datasets can be discarded. 

These two problems share common themes for future research: i) development of optimal strategies to train deep neural nets using TB-size datasets, e.g., using genetic algorithms; ii) development of statistical techniques to increase the sensitivity of neural nets to extract low signal-to-noise ratio signals from noisy time-series; iii) systematic exploration to elucidate why deep convolutional neural networks outperform machine learning classifiers, such as Random Forest, Support Vector Machine, k-Nearest Neighbors, Hidden Markov Model, Shallow Neural Networks, etc.,~\cite{DeepFiltering}, and explore whether this property is held in higher-order dimensional signal manifolds; and iv) assess whether deep learning point parameter estimation results are consistent with maximum likelihood Bayesian results, and thus useful as seeds to accelerate existing Bayesian formulations.

\section{State of the Art}

The key challenge to carry out real-time regression analysis is related to the fact that most deep neural network algorithms output point parameter estimation values. Ideally, we would like to also provide statistical information, as it is customarily done in Bayesian studies. Recent work has started to shed light in this direction~\cite{bayesian:171100165}. Being able to carry out real-time regression with deep neural networks that provide statistical information would be a remarkable achievement with far reaching consequences. 

To accomplish this work, we would have to be able to generalize to higher-dimensional signal manifolds the work we introduced in~\cite{DeepFiltering,DNNRealNoise}. To tackle this problem, it will be necessary to quantify whether the parameter space can be compactified, thereby removing parameter space degeneracies and accelerating the training time and hyperparameter optimization of neural nets. It will also be necessary to assess what type of neural nets are optimal for the problem at hand, i.e., we have found that recurrent neural nets are ideal for de-noising~\cite{Denoising} time-series, whether deep convolutional neural nets are optimal for regression and classification~\cite{DeepFiltering,DNNRealNoise}. 

In the gravitational wave detection scenario, this would imply that a single deep learning algorithm, running on a dedicated inference GPU, would suffice to process the lightweight data (2MB/second) that is generated in low latency by gravitational wave detectors. Similarly, if a similar framework is applied to process images, and extract specific signatures embedded in noise, such as the images to be generated by LSST~\cite{DeepTransferLearning,transient:171001422S} , then both time-series data and images could be post-processed simultaneously in real-time, facilitating the observation of astrophysical phenomena using multimessenger astronomy, i.e., contemporaneous observations with gravitational waves, light, neutrinos and cosmic rays. 

\section{Maturity and uniqueness}

Deep learning is uniquely posed to overcome what is known as the curse of dimensionality~\cite{DL-Book,Scaling}, since it is known to be highly scalable. This intrinsic ability of DNNs to take advantage of large datasets is a unique feature to enable classification and regression analyses over a higher dimensional parameter-space that is beyond the reach of existing algorithms. 

Furthermore, DNNs are excellent at generalizing or extrapolating to new data. In the context of gravitational wave astronomy, our preliminary results indicates that DNNs, trained with only signals from a 2-dimensional parameter space were able to detect and reconstruct the parameters of signals that span up to a 4-dimensional signal manifolds, and which currently may go unnoticed with established detection algorithms~\cite{Huerta:2017a,Tiwari:2016,Huerta:2014,Huerta:2013a}. With existing computational resources on supercomputers, such as Blue Waters, we estimate that it is feasible to train DNNs that target a 9D parameter space within a few weeks.

Furthermore, DNN algorithms requires minimal pre-processing. CNNs are capable of automatically learning to perform band-pass filtering on raw time-series inputs~\cite{DCNN-Raw}, and that they are excellent at suppressing highly non-stationary colored noise~\cite{SpeechEnhancement} especially when incorporating real-time noise characteristics~\cite{NoiseCues}. This suggests that manually devised pre-processing and whitening steps may be eliminated and raw data can be fed to DNNs. This would be particularly advantageous since it is known that Fourier transforms are the bottlenecks of matched-filtering based algorithms.

\section{Novelty} 

The deep learning algorithms we pioneered in~\cite{DeepFiltering,DNNRealNoise} constitute the first demonstration that deep convolutional neural networks can be applied for real-time classification and regression of weak signals embedded in non-stationary and non-Gaussian noise. It is also the first time that DNN were shown to be to exhibit features similar to Gaussian Process Regression~\cite{2003itil.book.....M,gpr:2016PhRvD,moore:2014gpr}, and to \textit{generalize} to signals beyond the templates used for training. Furthermore, our DNNs can be evaluated faster than real-time with a single CPU, and very intensive searches over a broader range of signals can be easily carried out with one dedicated GPU. These results have sparked a keen interest in the gravitational wave community, and have led to a plethora of independent studies within the gravitational wave physics and computer science community.





\section*{Acknowledgement}
This research is part of the Blue Waters sustained-petascale computing project, which is supported by the National Science Foundation (awards OCI-0725070 and ACI-1238993) and the State of Illinois. Blue Waters is a joint effort of the University of Illinois at Urbana-Champaign and its National Center for Supercomputing Applications. 

\bibliographystyle{ieeetr}
\bibliography{referencesfinal}

\end{document}